\documentclass[11pt,a4paper]{article}
\usepackage[hyperref]{emnlp-ijcnlp-2019}
\usepackage{times}
\usepackage{quoting}
\usepackage{latexsym}
\usepackage{graphicx}

\newcommand\B{\rule[-3ex]{0pt}{0pt}} 

\newcommand{\eat}[1]{}

\mathchardef\mhyphen="2D
\newenvironment{enu}{                   
     \parskip 0cm \begin{list}{}{\parsep 0cm \itemsep 0cm \topsep 0cm}}{
       \end{list}} 
\newenvironment{des}{                 
     \parskip 0cm \begin{list}{}{\parsep 0cm \itemsep 0cm \topsep 0cm}}{
       \end{list}} 
\newcommand{\QuaRTz}{\textsc{QuaRTz}}
\newcommand{\nquestions}{3864}
\newenvironment{myquote}{                   
  \parskip 0mm \begin{quoting}[vskip=0mm,leftmargin=3mm]}{
\end{quoting}}
\newenvironment{mycentering}
 {\parskip=0pt\par\nopagebreak\centering}
 {\par\noindent\ignorespacesafterend}

\newcommand{\red}[1]{\textcolor{red}{#1}}
\newcommand{\blue}[1]{\textcolor{blue}{#1}}
\newcommand{\orange}[1]{\textcolor{orange}{#1}}

\newcommand{\teal}[1]{\textcolor{teal}{#1}}

\usepackage{url}

\aclfinalcopy 

\title{\QuaRTz: An Open-Domain Dataset of Qualitative Relationship Questions}

\author{ 
Oyvind Tafjord, Matt Gardner, Kevin Lin, Peter Clark \\
        Allen Institute for Artificial Intelligence, Seattle, WA \\
        {\tt \{oyvindt,mattg,kevinl peterc\}@allenai.org}
        }
        
\date{}

\begin{document}
\maketitle
\begin{abstract}
We introduce the first open-domain dataset, called \QuaRTz, for reasoning about textual qualitative relationships.
\QuaRTz~contains general qualitative statements, e.g., ``A sunscreen with a higher SPF protects the skin longer.'',
twinned with \nquestions~crowdsourced situated questions, e.g., ``Billy is wearing sunscreen with a lower SPF than Lucy. Who
will be best protected from the sun?'', plus annotations of the properties being compared.
Unlike previous datasets, the general knowledge is textual and not tied to a fixed set of
relationships, and tests a system's ability to comprehend and apply textual qualitative knowledge in a
novel setting. We find state-of-the-art results are substantially (20\%) below human performance, presenting an
open challenge to the NLP community.
\end{abstract}

\section{Introduction}

Understanding and applying qualitative knowledge is a fundamental facet of intelligence.
For example, we may read that exercise improves health, and thus decide to spend more
time at the gym; or that larger cars cause more pollution, and thus decide to buy a smaller
car to be environmentally sensitive. These skills require understanding the
underlying qualitative relationships, and being able to apply them in specific
contexts.

\begin{figure}[t]
  \centerline{
\fbox{%
  \parbox{1\columnwidth}{\small
    {\bf Q:} If Mona lives in a city that begins producing a \blue{greater} \red{amount of pollutants}, what happens to the \teal{air quality} in that city?
    (A) increases (B) \orange{decreases} {\bf [correct]}  \vspace{1mm} \\
{\bf K:} \blue{More} \red{pollutants} mean \orange{poorer} \teal{air quality}. 
}}} {\small
\hspace*{0.5cm}           {\bf Annotations:} \\
\hspace*{0.5cm}           {\bf Q:} [{\bf MORE}, \blue{"greater"}, \red{"amount of pollutants"}] \\
\hspace*{1.0cm} $\rightarrow$ (A) [{\bf MORE}, \orange{"increases"}, \teal{"air quality"}] \\
\hspace*{1.5cm} (B) [{\bf LESS}, \orange{"decreases"}, \teal{"air quality"}] \\
\hspace*{0.5cm}   {\bf K:} [{\bf MORE}, \blue{"more"}, \red{"pollutants"}] \\
\hspace*{1.5cm} $\leftrightarrow$ [{\bf LESS}, \orange{"poorer"}, \teal{"air quality"}]
                  }
                 \caption{\QuaRTz~contains situated qualitative questions, each paired with a gold background knowledge sentence and qualitative annotations.
                   \label{example} }
                 \vspace{-3mm}
\end{figure}

\eat{
Here's the non-colorful version
\begin{figure}[t]
  \centerline{
\fbox{%
  \parbox{1\columnwidth}{\small
    {\bf Q:} If Mona lives in a city that begins producing a greater amount of pollutants, what happens to the air quality in that city?
    (A) increases (B) decreases {\bf [correct]}  \vspace{1mm} \\
{\bf K:} More pollutants mean poorer air quality. 
}}} {\small
\hspace*{0.5cm}           {\bf Annotations:} \\
\hspace*{0.5cm}           {\bf Q:} [\blue{MORE}, "greater", "amount of pollutants"] \\
\hspace*{1.0cm} $\rightarrow$ (A) [\blue{MORE}, "increases", "air quality"] \\
\hspace*{1.5cm} (B) [\blue{LESS}, "decreases", "air quality"] \\
\hspace*{0.5cm}   {\bf K:} [\blue{MORE}, "more", "pollutants"] \\
\hspace*{1.5cm} $\leftrightarrow$ [\blue{LESS}, "poorer", "air quality"]
                  }
                 \caption{\QuaRTz~contains situated qualitative questions, each paired with a gold background knowledge sentence and qualitative annotations.
                   \label{example}. }
                 \vspace{-3mm}
\end{figure}
}
To promote research in this direction, we present the first {\it open-domain}
dataset of qualitative relationship questions, called \QuaRTz~(``Qualitative Relationship Test set'')\footnote{Available at http://data.allenai.org/quartz/}. Unlike earlier
work in qualitative reasoning, e.g., \cite{quarel}, the dataset is not restricted to
a small, fixed set of relationships.
Each question $Q_i$ (2-way multiple choice) is grounded in a
particular situation, and is paired with a sentence $K_i$
expressing the general qualitative knowledge needed to answer it. $Q_i$ and $K_i$ are also
annotated with the properties being compared (Figure~\ref{example}).
The property annotations serve as supervision for a potential semantic parsing based approach.
The overall task is to answer the $Q_i$ given the corpus $K = \{K_i\}$.

We test several state-of-the-art (BERT-based) models and find that
they are still substantially (20\%) below human performance.
Our contributions are thus 
(1) the dataset, containing \nquestions~richly annotated questions plus a background corpus of 400 qualitative knowledge sentences; and
(2) an analysis of the dataset, performance of BERT-based models, and a catalog of the challenges it poses, pointing the way towards solutions.

\begin{table*}
  {\small 
    \begin{tabular}{|lp{6in}|} \hline

      \multicolumn{2}{|l|}{\bf Differing Comparatives:} \\

      $Q_1$ & Jan is comparing stars, specifically a small star and the \red{larger} Sun. Given the size of each, Jan can tell that the Sun puts out \blue{heat that is (A) greater} (B) lesser \\ 
       $K_1$ & \red{Bigger} stars produce more energy, so their surfaces are \blue{hotter}. \B \\
      \multicolumn{2}{|l|}{\bf Discrete Property Values:} \\
      $Q_2$ & What happens to a \blue{light} car when it has the same power as a \blue{heavy} car? (A) accelerates faster (B) accelerates slower \\
      $K_2$ & The \blue{smaller its mass} is, the greater its acceleration for a given amount of force. \B \\

      \multicolumn{2}{|l|}{\bf Numerical Property Values:} \\
      $Q_3$ & Will found water from a source \blue{10m from shore}. Eric found water from a source \red{2m from shore}. Whose water likely contains the least nutrients? (A) Will's (B) Eric's       \\
      $K_3$ & Most nutrients are washed into ocean water from land. Therefore, water \red{closer to shore} tends to have more nutrients.	\B \\


      \multicolumn{2}{|l|}{\bf Commonsense Knowledge:} \\
      $Q_4$ & Compared to a box of \blue{bricks} a box of \red{feathers} would be (A) lighter (B) heavier       \\
      $K_4$ & A given volume of a \blue{denser substance} is heavier than the same volume of a \red{less dense} substance. \B \\

      \multicolumn{2}{|l|}{\bf Multiple Entities (``Worlds''):} \\
      $Q_5$ & \red{Jimbo} liked to work out, while \blue{James} never did. Which person would have weaker muscles? (A) \red{Jimbo} (B) \blue{James}  \\    
      $K_5$ & Muscles that are exercised are bigger and stronger than muscles that are not exercised. \B \\

      \multicolumn{2}{|l|}{\bf Complex Stories:} \\
      $Q_6$ & NASA has sent an unmanned probe to survey a distant solar system with four planets. \red{Planet Zorb is farthest from the sun of this solar system, Planet Krakatoa is second farthest}, Planet Beanbag is third, and Krypton is the closest. The probe visits the planets in order, first Zorb, then Krakatoa, then Beanbag and finally Krypton. Did the probe have to \blue{fly farther in its trip} from (A) \red{Zorb to Krakatoa} or (B) from Beanbag to Krypton? \\
      $K_6$ & In general, the \red{farther away from the Sun}, the \blue{greater the distance} from one planets orbit to the next. \\
      \hline
  \end{tabular}
  }
  \caption{Examples of crowdsourced questions Q and corpus knowledge K in \QuaRTz, illustrating phenomena. \label{examples}}
  \vspace{-4mm}
\end{table*}

\section{Related Work}

Despite rapid progress in general question-answering (QA), e.g., \cite{clark2017simple},
and formal models for qualitative reasoning (QR), e.g., \cite{Forbus1984QualitativePT,weld2013readings},
there has been little work on reasoning with {\it textual} qualitative knowledge,
and no dataset available in this area. Although many datasets include
a few qualitative questions, e.g., \cite{yang2018hotpotqa,Clark2018ThinkYH},
the only one directly probing QR is QuaRel \cite{quarel}. However, although QuaRel contains 2700
qualitative questions, its underlying qualitative knowledge was specified formally,
using a small, fixed ontology of 19 properties. As a result, systems trained on QuaRel
are limited to only questions about those properties.
Likewise, although the QR community has performed some work on extracting
qualitative models from text, e.g., \cite{mcfate2014using,mcfate2016scaling},
and interpreting questions about identifying qualitative processes,
e.g., \cite{crouse2018learning}, there is no dataset
available for the NLP community to study textual qualitative reasoning.
\QuaRTz~addresses this need.

\section{The Task}

Examples of QuaRTz questions $Q_{i}$ are shown in Table~\ref{examples}, along with
a sentence $K_{i}$ expressing the relevant qualitative relationship.
The \QuaRTz~task is to answer the questions given a small (400 sentence)
corpus $K$ of general qualitative relationship sentences. Questions are crowdsourced,
and the sentences $K_{i}$ were collected from a larger corpus, described shortly.

Note that the task involves substantially more than matching intensifiers
(more/greater/...) between $Q_{i}$ and $K_{i}$. Answers also require
some qualitative reasoning, e.g., if the intensifiers are inverted
in the question, and entity tracking, to keep track of which entity
an intensifier applies to. For example, consider the qualitative sentence and three
questions (correct answers bolded):

\eat{\small
  \noindent
$K_{n}$: People with greater height are stronger. \\
$Q_{n}$: Sue is taller than Joe so Sue is (A) {\bf stronger} (B) weaker \\
$Q_{n}'$: Sue is shorter than Joe so Sue is (A) stronger (B) {\bf weaker} \\
$Q_{n}''$: Sue is shorter than Joe so Joe is (A) {\bf stronger} (B) weaker
}

\begin{des}
\item[$K_{n}$:] People with greater height are stronger. 
\item[$Q_{n}$:] Sue is taller than Joe so Sue is (A) {\bf stronger} (B) weaker 
\item[$Q_{n}'$:] Sue is shorter than Joe so Sue is (A) stronger (B) {\bf weaker} 
\item[$Q_{n}''$:] Sue is shorter than Joe so Joe is (A) {\bf stronger} (B) weaker
  \end{des}

\noindent
$Q_{n}'$ requires reasoning about intensifers that are flipped with respect to $K$ (shorter $\rightarrow$ weaker),
and $Q_{n}''$ requires entities be tracked correctly (asking about Sue or Joe changes the answer).

\section{Dataset Collection \label{dataset}}

\QuaRTz~was constructed as follows. First, 400 sentences\footnote{
    In a few cases, a short paragraph rather than sentence was selected, where surrounding context was needed
    to make sense of the qualitative statement.}
    expressing general qualitative relations were manually extracted by the authors from a large corpus using
    keyword search (``increase'', ``faster'', etc.).
    Examples ($K_i$) are in Table~\ref{examples}.

Second, crowdworkers were shown a seed sentence $K_i$, and asked to
annotate the two properties being compared using the template below, illustrated
using $K_2$ from Table~\ref{examples}:

\begin{myquote}
\noindent
"The smaller its mass is, the greater its acceleration for a given amount of force."
\end{myquote}

\vspace{2mm}
\centerline{\includegraphics[width=0.9\columnwidth]{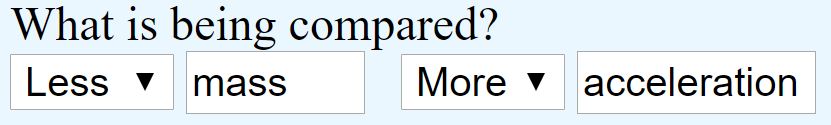}}
\vspace{1mm}

\noindent
They were then asked to author a situated, 2-way multiple-choice (MC) question
that tested this relationship,
guided by multiple illustrations.
Examples of their questions ($Q_i$) are in Table~\ref{examples}.

 Third, a second set of workers was shown an authored question,
 asked to validate its answer and quality,
 and asked to annotate how the properties of $K_i$ identified earlier
 were expressed. To do this, they filled a second template,
 illustrated for $Q_2$:

 \vspace{2mm}
\centerline{\includegraphics[width=1\columnwidth]{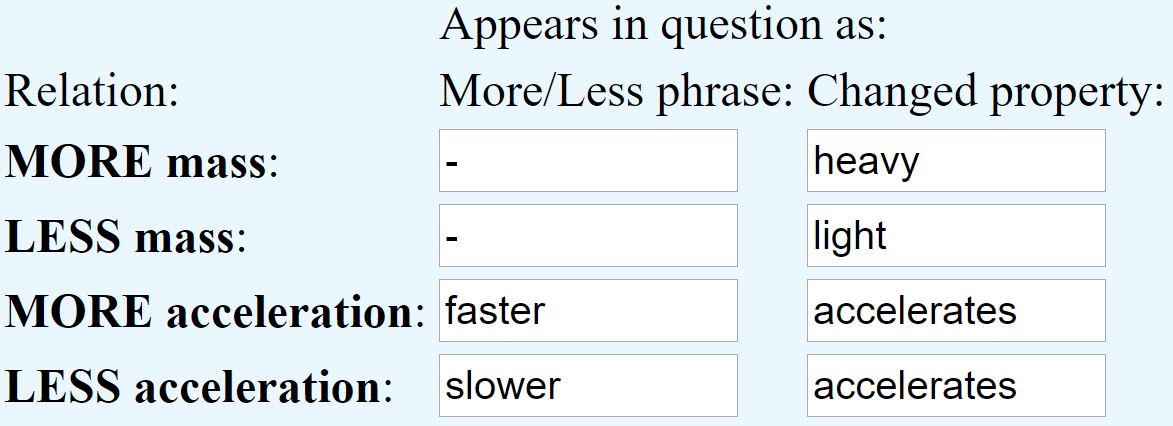}}
\vspace{1mm}

Finally these workers were asked to generate a new question by ``flipping'' the original so the answer switched.
Flipping means inverting comparatives (e.g., ``more'' $\rightarrow$ ``less''), values, and other edits as needed to change the answer, 
e.g., \\
K: {\it More rain causes damper surfaces.} \\
Q:{\it More rain causes (A) wetter land (B) drier land}\\
Q-flipped: {\it {\bf Less} rain causes (A) wetter land (B)} \\
\hspace*{1.6cm} {\it drier land}\\
Flipped questions are created to counteract the tendency of workers to use the
same comparison direction (e.g., ``more'') in their question as in
the seed sentence $K_i$, potentially answerable by simply matching
$Q_i$ and $K_i$. Flipped questions are more challenging as they demand more qualitative reasoning (Section~\ref{qualitative-reasoning}).

\eat{
For example, for the question in Figure~\ref{example}:
\\
Q: more sun causes (A) drier grass (B) wetter grass\\
K: ``More pollution causes worse air.''\\
Q-flip: less pollution means (A) better air (B) worse air

{\it  If Mona lives in a city that begins producing a {\bf smaller} amount of pollutants, what happens to the air quality...? (A) increases (B) decreases}
Flipped questions are created to counteract the tendency of workers to use the
same comparison direction (e.g., ``bigger'') in their question as in
the seed sentence $K_i$. By flipping, we also obtain questions using the
opposite comparison direction (e.g., ``smaller'') to balance the dataset.
}

Questions marked by workers as poor quality were reviewed by
the authors and rejected/modified as appropriate.
The dataset was then split into train/dev/test partitions such that questions from the same seed $K_i$ were
all in the same partition. Statistics are in Table~\ref{statistics}.


To determine if the questions are correct and answerable given the general
knowledge, a human baseline was computed. Three annotators independently
answered a random sample of 100 questions given the supporting sentence $K_i$ for each.
The mean score was 95.0\%.

\begin{table}
  \begin{mycentering}
  {\small
  \begin{tabular}{|l|l|} \hline
    \# questions $Q_i$ & \nquestions \\
    \hspace*{0.5cm} flip/no flip & 1932/1932 \\
    \hspace*{0.5cm} positive/negative qualitative & \\
    \hspace*{1.5cm}{influence (QR+/QR-)} & 2772/1092 \\
    \hspace*{0.5cm} train/dev/test & 2696/384/784 \\
    av $Q_i$ length (sents) min/avg/max  & 1/1.5/6 \\
    av $K_i$ length (sents) min/avg/max & 1/1.1/4 \\ \hline
  \end{tabular}
  }
  \end{mycentering}  
  \caption{Statistics of \QuaRTz. \label{statistics}}
\end{table}

\section{Models}
\vspace{-1mm}
The \QuaRTz~task is to answer the questions given the corpus $K$ of qualitative background
knowledge. We also examine a ``no knowledge'' (questions only) task and a ``perfect knowledge'' task
(each question paired with the qualitative sentence $K_i$ it was based on).
We report results using two baselines and several strong models built with
BERT-large \cite{devlin2018bert} as follows:

\noindent
1. {\bf Random:} always 50\% (2-way MC).

\noindent
2. {\bf BERT-Sci:} BERT fine-tuned on a large, general set of science questions \cite{Clark2018ThinkYH}.

\noindent
3. {\bf BERT (IR):} This model performs the full task. 
First, a sentence $K_i$ is retrieved from $K$ using $Q_i$ as a search query. This is then
  supplied to BERT as {\it [CLS] $K_i$ [SEP] question-stem [SEP]
  answer-option [SEP]} for each option.
The [CLS] output token is projected to a single logit and fed through a softmax layer across answer options, using cross entropy loss, the highest being selected.
This model is fine-tuned using \QuaRTz~(only).

\noindent
4. {\bf BERT (IR upper bound):} Same, but using the ideal (annotated) $K_i$ rather than retrieved $K_i$. 

\noindent
5. {\bf BERT-PFT (no knowledge):} BERT first fine-tuned (``pre-fine-tuned'') on the RACE dataset \cite{lai2017race, Sun2018ImprovingMR},
  and then fine-tuned on \QuaRTz~(questions only, no $K$, both train and test).
  Questions are supplied as {\it [CLS] question-stem [SEP] answer-option [SEP]}.

\noindent
6. {\bf BERT-PFT (IR):} Same as BERT (IR), except starting with the pre-fine-tuned BERT.

\eat{
\item[3.] {\bf BERT (no knowledge):} BERT fine-tuned on \QuaRTz~questions only. Questions are 
  supplied to BERT as {\it [CLS] question-stem [SEP] answer-option [SEP]} for each answer-option. As is common, the [CLS] output token is projected to a single logit and fed through a softmax layer across answer options, using cross entropy loss. 
\item[4.] {\bf BERT-PFT (no knowledge):} BERT first fine-tuned (``pre-fine-tuned'') on the RACE dataset \cite{lai2017race, Sun2018ImprovingMR},
  and then fine-tuned on \QuaRTz~(questions only).
\item[5.] {\bf BERT-PFT (IR):} This model performs the full task. First, a sentence $K_i$ is
  retrieved from $K$ using $Q_i$ as a search query. This is then
  supplied to BERT as {\it [CLS] $K_i$ [SEP] question-stem [SEP] answer-option [SEP]} for
  each option.
\item[6.] {\bf BERT-PFT (IR upper bound):} Same except using the $K_i$ paired with $Q_i$,
  representing the performance with ideal IR.
  }
 \noindent All models were implemented using AllenNLP \cite{gardner2018allennlp}.

\section{Results}
\vspace{-1mm}
\begin{table}
{\small
\setlength\tabcolsep{6pt}    	
  \begin{tabular}{|l||l|ll|} \hline
\multicolumn{1}{|c||}{\bf Questions$\rightarrow$} & {\bf All} & {\bf No-flip} & {\bf Flip}  \\
  {\bf Model $\downarrow$} 	& {\bf Test}	& {\bf only} 	&  {\bf only}	\\ \hline
  {\bf Baselines:}   & & & \\
Random & 50.0 & 50.0 & 50.0 \\
BERT-Sci & 54.6	& 76.0	& 33.2	\\ 
\hline
  {\bf Models:} & & &  \\
BERT (IR) & 64.4 & 66.3 & 62.5  \\
(BERT IR upper bound) & (67.7) & (68.1) & (67.3)  \\
BERT-PFT (no knowledge) 	& 68.8	 & 70.4	 & 67.1	 \\
BERT-PFT (IR) 	  	& {\bf 73.7}	 & 77.3	 & 70.2	 \\
(BERT-PFT IR upper bound)& (79.8)	 & (82.1)	 & (77.6)	  \\ \hline
Human & 95.0 & & \\ \hline
  \end{tabular}
  }
  \caption{Performance of various models on \QuaRTz. \label{results}}
  \vspace{-3mm}
\end{table}

\eat{
\begin{table}
{\small
\setlength\tabcolsep{2pt}    	
  \begin{tabular}{|l||l|ll|ll|} \hline
\multicolumn{1}{|c||}{\bf Questions$\rightarrow$} & {\bf All} & {\bf No-flip} & {\bf Flip} & {\bf QR+} & {\bf QR-} \\
  {\bf Model $\downarrow$} 	& {\bf Test}	& {\bf only} 	&  {\bf only}	& {\bf only} 	& {\bf only}	\\ \hline
\multicolumn{1}{|r||}{Number of Qns:}	 & 784	 & 392	 & 392	 & 542	 & 242	 \\
  {\bf Baselines:}   & & & & & \\
Random & 50.0 & 50.0 & 50.0 & 50.0 & 50.0 \\
BERT-Sci & 54.6	& 76.0	& 33.2	& 57.0	& 49.2	\\
  {\bf Models:} & & & & & \\
BERT (IR) & 64.4 & 66.3 & 62.5 & 67.5 & 57.4 \\
(BERT IR upper bound) & (67.7) & (68.1) & (67.3) & (72.7) & (56.6) \\
BERT-PFT (no knowledge) 	& 68.8	 & 70.4	 & 67.1	 & 77.7	 & 48.8	 \\
BERT-PFT (IR) 	  	& {\bf 73.7}	 & 77.3	 & 70.2	 & 78.6	 & 62.8	 \\
(BERT-PFT IR upper bound)& (79.8)	 & (82.1)	 & (77.6)	 & (82.5)	 & (74.0) \\ \hline
Human & 95.0 & & & & \\ \hline
  \end{tabular}
  }
  \caption{Performance of various models on \QuaRTz. \label{results}}
  \vspace{-3mm}
\end{table}
}

The results are shown in Table~\ref{results}, and provide insights into both the data and the models:

\noindent
1. {\bf The dataset is hard.} Our best model, BERT-PFT (IR), scores only 73.7, over 20 points behind human performance (95.0),
suggesting there are significant linguistic and semantic challenges to overcome (Section~\ref{analysis}).

\noindent
2. {\bf A general science-trained QA system has not learned this style of reasoning.} BERT-Sci only scores 54.6, just a little above random (50.0).

\noindent
3. {\bf Pre-Fine-Tuning is important.} Fine-tuning only on \QuaRTz~does signficantly worse (64.4) than pre-fine-tuning on RACE before fine-tuning on \QuaRTz~(73.7). Pre-fine-tuning appears to teach BERT something about multiple choice questions in general, helping it more effectively fine-tune on \QuaRTz.

\noindent
4. {\bf BERT already ``knows'' some qualitative knowledge.} Interestingly, BERT-PFT (no knowledge) scores 68.8, significantly above random, suggesting that BERT already ``knows'' some kind of qualitative knowledge. To rule out annotation artifacts, we
we experimented with balancing the distributions of positive and negative influences, and different train/test splits to ensure no topical overlap between train and test, but the scores remained consistent.

\noindent
5. {\bf BERT can apply general qualitative knowledge to QA, but only partially.} The model for the full task, BERT-PFT (IR) outperforms the no knowledge version (73.7, vs. 68.8), but still over 20 points below human performance. Even given the ideal knowledge (IR upper bound), it is still substantially behind (at 79.8) human performance. This suggests more sophisticated ways of training and/or reasoning with the knowledge are needed.

\section{Discussion and Analysis \label{analysis}}
\vspace{-2mm}

\subsection{Qualitative Reasoning \label{qualitative-reasoning}}

Can models learn qualitative reasoning from \QuaRTz? While \QuaRTz~questions do not require chaining,
50\% involve ``flipping'' a qualitative relationship (e.g., K: ``more X $\rightarrow$ more Y'', Q: ``Does less X $\rightarrow$ less Y?'').
Training on just the original crowdworkers' questions, where they chose to flip the knowledge only 10\% of the time,
resulted in poor (less than random) performance on all the flipped questions. However, training on full \QuaRTz,
where no-flip and flip were balanced, resulted in similar score for both types of question,
suggesting that such a reasoning capability can indeed be learned.

\begin{table}[t]
  {\small
    \begin{mycentering}    
\begin{tabular}{|c|} \hline
  \underline{\bf Comparatives:} \\
  ``warmer''  $\leftrightarrow$ ``increase temperature'' \\
``more difficult''  $\leftrightarrow$ ``slower'' \\
``need more time''  $\leftrightarrow$ ``have lesser amount'' \\
``decreased distance''  $\leftrightarrow$ ``hugged'' \\
``cost increases''  $\leftrightarrow$ ``more costly'' \\
``increase mass''  $\leftrightarrow$ ``add extra'' \\
  ``more tightly packed''  $\leftrightarrow$ ``add more'' \B \\
  \underline{\bf Commonsense Knowledge:} \\  
``more land development''  $\leftrightarrow$ ``city grow larger'' \\
  ``not moving''  $\leftrightarrow$ ``sits on the sidelines'' \\
    ``caught early'' $\leftrightarrow$ `sooner treated'' \\
``lets more light in''  $\leftrightarrow$ ``get a better picture'' \\
``stronger electrostatic force''  $\leftrightarrow$ ``hairs stand up more'' \\
``less air pressure''  $\leftrightarrow$ ``more difficult to breathe'' \\
  ``more photosynthesis''  $\leftrightarrow$ ``increase sunlight'' \B \\
  \underline{\bf Discrete Values:} \\    
``stronger acid'' $\leftrightarrow$ ``vinegar'' vs. ``tap water'' \\
``more energy'' $\leftrightarrow$ ``ripple'' vs. ``tidal wave'' \\
``closer to Earth'' $\leftrightarrow$ ``ball on Earth'' vs. ``ball in space'' \\
``mass'' $\leftrightarrow$ ``baseball'' vs. ``basketball'' \\
``rougher'' $\leftrightarrow$ ``notebook paper'' vs. ``sandpaper'' \\
``heavier'' $\leftrightarrow$ ``small wagon'' vs. ``eighteen wheeler'' \\ \hline
\end{tabular}
\end{mycentering}
  }
  \caption{Examples of linguistic and semantic gaps between knowledge $K_i$ (left) and question $Q_i$ (right).
    A system needs to bridge such gaps for high performance. \label{gaps}}
  \vspace{-8mm}
\end{table}

\subsection{Linguistic Phenomena}

From a detailed analysis of 100 randomly sampled questions, the large majority (86\%) involved the (overlapping)
linguistic and semantic phenomena below, and illustrated in Tables~\ref{examples} and~\ref{gaps}:
\begin{enu}
\item[1.] {\bf Differing comparative expressions} ($\approx$68\%) between $K_i$ and $Q_i$ occur in the majority of questions, e.g.,
  \begin{mycentering}
    {\it  ``increased altitude'' $\leftrightarrow$ ``higher'' }
    \end{mycentering}
\item[2.] {\bf Indirection and Commonsense knowledge} ($\approx$35\%) is needed for about 1/3 of the
  questions to relate $K$ and $Q$, e.g.,\\
{\it ``higher temperatures'' $\leftrightarrow$ ``A/C unit broken''}
\item[3.] {\bf Multiple Worlds} ($\approx$26\%): 
  1/4 of the questions explicitly mention {\it both} situations being compared, e.g., $Q_1$ in Table~\ref{examples}.
  Such questions are known to be difficult because models can easily confuse the two situations \cite{quarel}.
\item[4.] {\bf Numerical property values} ($\approx$11\%) require numeric comparison to identify the qualitative
  relationship, e.g., that ``60 years'' is older than ``30 years''.
\item[5.] {\bf Discrete property values} ($\approx$7\%), often require commonsense to compare,
  e.g., that a ``melon'' is larger than an ``orange''.
\item[6.] {\bf Stories} ($\approx$15\%): 15\% of the questions are 3 or more sentences long, making comprehension more challenging.
\end{enu}
This analysis illustrates the richness of linguistic and semantic phenomena in \QuaRTz.
  
\subsection{Use of the Annotations}

\QuaRTz~includes a rich set of annotations on all the knowledge sentences and questions,
marking the properties being compared, and the linguistic and semantic comparatives employed (Figure~\ref{example}).
This provides a laboratory for exploring semantic parsing approaches, e.g., \cite{berant2013semantic,Krishnamurthy2017NeuralSP},
where the underlying qualitative comparisons are extracted and can be reasoned about.

\section{Conclusion}

Understanding and applying textual qualitative knowledge is an important skill for question-answering,
but has received limited attention, in part due the lack of a broad-coverage dataset
to study the task. \QuaRTz~aims to fill this gap by providing the first {\it open-domain}
dataset of qualitative relationship questions, along with the requisite
qualitative knowledge and a rich set of annotations.
Specifically, QuaRTz 
removes the requirement, present in all previous qualitative reasoning work, that a fixed set of qualitative relationships be formally pre-specified. Instead, QuaRTz tests the ability of a system to find and apply an arbitrary relationship on the fly to answer a question, including when simple reasoning (arguments, polarities) is required.


As the \QuaRTz~task involves using a general corpus $K$ of textual qualitative
knowledge, a high-performing system would be close to a fully general
system where $K$ was much larger (e.g., the Web or a filtered subset),
encompassing many more qualitative relationships, and able to 
answer arbitrary questions of this kind.
Scaling further would thus require more sophisticated retrieval over a larger corpus, and (sometimes)
chaining across influences, when a direct connection was not described in the corpus.
\QuaRTz~thus provides a dataset towards this end, allowing controlled
experiments while still covering a substantial number of textual relations in an open setting.
QuaRTz is available at http://data.allenai.org/quartz/.

\eat{
\section{Conclusion}
\vspace{-2mm}
Understanding and applying textual qualitative knowledge is an important skill for question-answering,
but has received limited attention, in part due the lack of a broad-coverage dataset
to study the task. \QuaRTz~aims to fill this gap by providing the first {\it open-domain}
dataset of qualitative relationship questions, along with the requisite
qualitative knowledge and a rich set of annotations.
As the \QuaRTz~task involves using a general corpus $K$ of textual qualitative
knowledge, a high-performing system would thus be close to a fully general
system where $K$ was much larger (e.g., the Web or a filtered subset),
encompassing many more qualitative relationships, and able to 
answer arbitrary questions of this kind.
We hope that \QuaRTz~provides a testbed to help facilitate such an advance.
}

\subsection*{Acknowledgements}
We are grateful to the AllenNLP and Beaker teams at AI2, and for the insightful discussions with other Aristo team members. 
Computations on beaker.org were supported in part by credits from Google Cloud.

\bibliography{main}

\begin{thebibliography}{15}
\expandafter\ifx\csname natexlab\endcsname\relax\def\natexlab#1{#1}\fi

\bibitem[{Berant et~al.(2013)Berant, Chou, Frostig, and
  Liang}]{berant2013semantic}
Jonathan Berant, Andrew Chou, Roy Frostig, and Percy Liang. 2013.
\newblock Semantic parsing on {Freebase} from question-answer pairs.
\newblock In \emph{EMNLP'13}.

\bibitem[{Clark and Gardner(2018)}]{clark2017simple}
Christopher Clark and Matt Gardner. 2018.
\newblock Simple and effective multi-paragraph reading comprehension.
\newblock In \emph{ACL}.

\bibitem[{Clark et~al.(2018)Clark, Cowhey, Etzioni, Khot, Sabharwal, Schoenick,
  and Tafjord}]{Clark2018ThinkYH}
Peter Clark, Isaac Cowhey, Oren Etzioni, Tushar Khot, Ashish Sabharwal, Carissa
  Schoenick, and Oyvind Tafjord. 2018.
\newblock Think you have solved question answering? {Try} {ARC}, the {AI2}
  reasoning challenge.
\newblock \emph{arXiv preprint arXiv:1803.05457}.

\bibitem[{Crouse et~al.(2018)Crouse, McFate, and Forbus}]{crouse2018learning}
M.~Crouse, C.~McFate, and K.~Forbus. 2018.
\newblock Learning to build qualitative scenario models from natural language.
\newblock In \emph{Proc. 31st Int. Workshop on Qualitative Reasoning (QR'18)}.

\bibitem[{Devlin et~al.(2019)Devlin, Chang, Lee, and
  Toutanova}]{devlin2018bert}
Jacob Devlin, Ming-Wei Chang, Kenton Lee, and Kristina Toutanova. 2019.
\newblock {BERT}: Pre-training of deep bidirectional transformers for language
  understanding.
\newblock In \emph{NAACL}.

\bibitem[{Forbus(1984)}]{Forbus1984QualitativePT}
Kenneth~D. Forbus. 1984.
\newblock Qualitative process theory.
\newblock \emph{Artificial Intelligence}, 24:85--168.

\bibitem[{Gardner et~al.(2018)Gardner, Grus, Neumann, Tafjord, Dasigi, Liu,
  Peters, Schmitz, and Zettlemoyer}]{gardner2018allennlp}
Matt Gardner, Joel Grus, Mark Neumann, Oyvind Tafjord, Pradeep Dasigi, Nelson
  Liu, Matthew Peters, Michael Schmitz, and Luke Zettlemoyer. 2018.
\newblock {AllenNLP}: A deep semantic natural language processing platform.
\newblock In \emph{NLP OSS Workshop at ACL}.
\newblock (arXiv:1803.07640).

\bibitem[{Krishnamurthy et~al.(2017)Krishnamurthy, Dasigi, and
  Gardner}]{Krishnamurthy2017NeuralSP}
Jayant Krishnamurthy, Pradeep Dasigi, and Matthew Gardner. 2017.
\newblock Neural semantic parsing with type constraints for semi-structured
  tables.
\newblock In \emph{EMNLP'17}.

\bibitem[{Lai et~al.(2017)Lai, Xie, Liu, Yang, and Hovy}]{lai2017race}
Guokun Lai, Qizhe Xie, Hanxiao Liu, Yiming Yang, and Eduard Hovy. 2017.
\newblock Race: Large-scale reading comprehension dataset from examinations.
\newblock In \emph{EMNLP}.

\bibitem[{McFate and Forbus(2016)}]{mcfate2016scaling}
Clifton McFate and Kenneth Forbus. 2016.
\newblock Scaling up linguistic processing of qualitative processes.
\newblock In \emph{Proc. 4th Ann. Conf. on Advances in Cognitive Systems}.

\bibitem[{McFate et~al.(2014)McFate, Forbus, and Hinrichs}]{mcfate2014using}
Clifton~James McFate, Kenneth~D Forbus, and Thomas~R Hinrichs. 2014.
\newblock Using narrative function to extract qualitative information from
  natural language texts.
\newblock In \emph{AAAI'14}.

\bibitem[{Sun et~al.(2019)Sun, Yu, Yu, and Cardie}]{Sun2018ImprovingMR}
Kai Sun, Dian Yu, Dong Yu, and Claire Cardie. 2019.
\newblock Improving machine reading comprehension with general reading
  strategies.
\newblock In \emph{NAACL}.

\bibitem[{Tafjord et~al.(2019)Tafjord, Clark, Gardner, Yih, and
  Sabharwal}]{quarel}
Oyvind Tafjord, Peter Clark, Matt Gardner, Wen-tau Yih, and Ashish Sabharwal.
  2019.
\newblock Quarel: A dataset and models for answering questions about
  qualitative relationships.
\newblock In \emph{{AAAI}}.

\bibitem[{Weld and De~Kleer(2013)}]{weld2013readings}
Daniel~S Weld and Johan De~Kleer. 2013.
\newblock \emph{Readings in qualitative reasoning about physical systems}.
\newblock Morgan Kaufmann.

\bibitem[{Yang et~al.(2018)Yang, Qi, Zhang, Bengio, Cohen, Salakhutdinov, and
  Manning}]{yang2018hotpotqa}
Zhilin Yang, Peng Qi, Saizheng Zhang, Yoshua Bengio, William~W Cohen, Ruslan
  Salakhutdinov, and Christopher~D Manning. 2018.
\newblock {HotpotQA}: A dataset for diverse, explainable multi-hop question
  answering.
\newblock In \emph{EMNLP}.

\end{thebibliography}
\bibliographystyle{./acl_natbib}

\end{document}